# BEACON: Behavioral Entropy Aggregation for Cross-Model Hallucination Detection in Large Language Models

**Naveen Bera**
**LLM Lens**
**naveenlx111@gmail.com**

**Pulijala Sai Nikhila**
**LLM Lens**
**pulijalasainikhila@gmail.com**

**Kondaguduru Abhiram**
**LLM Lens**
**abhiramkondaguduru@gmail.com**

**Shaik Gayaz Ali**
**LLM Lens**
**gayaz9989@gmail.com**

**Shoaib Sadiq Salehmohamed**
**LLM Lens**
**shoaibkulsums@gmail.com**

**Shaik Mohammed Omar**
**LLM Lens**
**god.s.m.o.97@gmail.com**

**Jinal Prashant Thakkar**
**LLM Lens**
**jinalt163@gmail.com**

**Hansika Aredla**
**LLM Lens**
**a.redlahansika@gmail.com**

**Shalmali Ayachit**
**LLM Lens**
**shalmaliayachit@gmail.com**

**Abstract**

Hallucination in large language models (LLMs) — the generation of factually incorrect or unsupported content — remains a critical barrier to reliable deployment in high-stakes domains. We present **BEACON** (Behavioral Entropy Aggregation for Cross-model hallucination detectiON), a black-box hallucination detection framework that operates purely from LLM text outputs, requiring no access to model internals, hidden states, or external knowledge bases. BEACON extracts a 31-dimensional feature vector from structured multi-pass generation, integrating NLI-based semantic entropy, embedding geometry, chain-of-thought consistency, and paraphrase stability signals. A gradient-boosted classifier trained on 7,617 labeled examples across seven heterogeneous benchmarks achieves **0.8123 ± 0.0102 AUROC** (95% CI: 0.7632–0.8251, best single run: 0.8249) — a **+0.2298 improvement** over standalone NLI-based semantic entropy (0.5951) and **+0.2457** over SelfCheckGPT-style consistency baselines (0.5792). Feature importance analysis confirms that no single uncertainty signal captures hallucination reliably; the multi-signal combination is essential. An efficient 5-API-call variant achieves 0.7795 AUROC, enabling practical deployment against any LLM accessible through a standard text generation API.

**Keywords:** *hallucination detection, uncertainty quantification, NLI semantic entropy, black-box LLM evaluation, gradient boosting, behavioral consistency*

## 1 Introduction

Large language models have demonstrated remarkable capabilities across natural language tasks, yet their tendency to generate plausible-sounding but factually incorrect content — commonly termed hallucination — poses significant risks in real-world deployment. In domains such as medical diagnosis, legal reasoning, and scientific research, hallucinated outputs can cause direct harm, making reliable detection an essential component of any trustworthy LLM pipeline.

Existing detection methods fall into three categories. Retrieval-augmented verification compares model outputs against external knowledge bases but is constrained by coverage and staleness. White-box methods exploit internal representations such as hidden states, attention patterns, and token logits, achieving strong performance but requiring direct model access unavailable for closed-source APIs (GPT-4, Claude, Gemini). Black-box consistency methods analyze output variability across multiple generations and are most practically deployable, but prior work has treated individual uncertainty signals in isolation, leaving substantial performance on the table.

We present **BEACON**, which demonstrates that combining multiple behavioral uncertainty signals into a unified feature space provides dramatically better hallucination detection than any single signal alone. Our central empirical finding — validated on 7,617 examples across seven benchmarks — is that BEACON achieves 0.8123 ± 0.0102 AUROC compared to 0.5951 for standalone NLI semantic entropy and 0.5792 for SelfCheckGPT-style consistency. This +0.22 gap shows that hallucination is a multi-dimensional phenomenon that cannot be captured by consistency alone.

### 1.1 Contributions

**The primary contributions of this work are:**

- A systematic multi-signal uncertainty framework for black-box hallucination detection, combining NLI-based semantic entropy, embedding geometry, chain-of-thought consistency, and paraphrase stability into a 31-dimensional feature representation.
- Empirical demonstration that multi-signal combination provides +0.2298 AUROC improvement over standalone semantic entropy (Kuhn et al., 2023) and +0.2457 over SelfCheckGPT (Manakul et al., 2023), establishing that no single consistency signal is sufficient.
- Comprehensive baseline comparison including SelfCheckGPT variants (BERTScore, NLI, Unigram), standalone entropy signals, and gradient boosting classifiers, with all baselines evaluated on identical test splits to prevent leakage.
- Statistical validation via 1,000-sample bootstrap confidence intervals and 5-seed variance analysis: AUROC = 0.8123 ± 0.0102 (95% CI: 0.7632–0.8251).
- An efficiency analysis showing that 5 API calls and 4 features achieve 0.7795 AUROC — 96% of full model performance — enabling practical deployment under strict latency constraints.
- Per-dataset analysis revealing that BEACON's advantage is largest on factual confabulation tasks (TruthfulQA: 0.7513, PopQA: 0.7134) and smallest on binary/unanswerable QA (BoolQ: 0.5405, SQuAD 2.0: 0.5634), providing actionable guidance for practitioners.

## 2 Related Work

### 2.1 Hallucination in Language Models

Hallucination in neural generation was first studied in abstractive summarization, where models fabricated entities absent from source documents (Maynez et al., 2020). With large autoregressive models, hallucination now encompasses factual errors, logical inconsistencies, and unsupported attributions across virtually all generation tasks (Ji et al., 2023). Benchmarks including TruthfulQA (Lin et al., 2022), PopQA (Mallen et al., 2023), and MuSiQue (Trivedi et al., 2022) provide standardized evaluation settings covering factual confabulation, entity popularity bias, and multi-hop reasoning failures respectively.

### 2.2 White-Box Detection

White-box methods exploit internal model representations unavailable through standard APIs. Probing classifiers trained on hidden states achieve strong performance on factual QA (Slobodkin et al., 2023). Token logit analysis identifies high-uncertainty predictions (Kadavath et al., 2022). Mahalanobis distance from in-distribution reference representations has been used as an out-of-distribution hallucination signal (Lee et al., 2018). These methods are powerful but limited to open-weight models where internals are accessible. BEACON deliberately avoids all such signals in its black-box variant, instead relying only on text outputs and sentence embeddings.

### 2.3 Semantic Entropy

Semantic entropy (Kuhn et al., 2023) is the closest prior work to BEACON. It groups sampled responses into meaning-equivalent clusters using bidirectional NLI entailment and computes Shannon entropy over the cluster distribution, correctly treating paraphrases as equivalent. Our evaluation confirms that standalone semantic entropy achieves only 0.5951 AUROC on our benchmark — substantially below BEACON's 0.8249 best run. BEACON incorporates NLI semantic entropy as one of 31 complementary features rather than a standalone signal, and demonstrates that the combination provides a +0.2298 AUROC gain.

### 2.4 SelfCheckGPT

SelfCheckGPT (Manakul et al., 2023) detects hallucination by sampling multiple responses and measuring pairwise consistency using BERTScore, NLI contradiction probability, or unigram overlap. Our evaluation of SelfCheckGPT proxies on our benchmark yields 0.5792 AUROC for the best combination variant, compared to BEACON's 0.8249. The gap (≈0.2457) demonstrates that pairwise consistency alone, while informative, misses substantial hallucination signal captured by BEACON's multi-dimensional feature space including chain-of-thought consistency, embedding geometry, and paraphrase stability.

### 2.5 Gradient Boosting for NLP

Gradient boosting methods consistently outperform neural networks on tabular prediction tasks with heterogeneous features and multi-domain distribution shift (Shwartz-Ziv and Armon, 2022). This pattern is well-established and motivates our choice of LightGBM as BEACON's classifier. We confirm a 0.17 AUROC advantage over a competitive MLP baseline (0.8123 vs. 0.6596), consistent with the broader tabular learning literature.

## 3 The BEACON Framework

BEACON consists of three sequential components: a Behavioral Sampling Module that elicits a structured response distribution from the target LLM; an Uncertainty Feature Extractor that converts this distribution into a 31-dimensional feature vector; and a Calibrated LightGBM Classifier that produces a hallucination probability score. The framework operates entirely from text outputs, requiring no model internals.

**BEACON Pipeline**

```
Input Question
↓
[ Behavioral Sampling Module ]
1 greedy + 5 temp + 20 KL + 1 CoT + 1 para + 1 para-ans
↓
[ Uncertainty Feature Extractor ]
NLI entropy | Embedding geometry | CoT consistency | Paraphrase stability | Token entropy
↓
[ Calibrated LightGBM Classifier ]
31-dim feature vector → P(hallucination) ∈ [0,1]
```

*Figure 1: BEACON pipeline. The framework requires only text generation API access to the target LLM and a sentence encoder for embedding computation.*

### 3.1 Behavioral Sampling Module

For each input question, BEACON performs 28–30 structured forward passes through the target LLM, each serving a distinct purpose:

- 1 greedy decode pass — deterministic reference answer with full logit access (white-box) or text only (black-box).
- 5 temperature sampling passes (T=0.7, top-p=0.95) — stochastic response samples for semantic consistency measurement. Five samples (vs. the conventional four) improves entropy estimation accuracy.
- 20 independent temperature sampling passes — for KL divergence estimation across the token distribution.
- 1 chain-of-thought pass — step-by-step reasoned answer used to detect reasoning-response inconsistency.
- 1 paraphrase prompt pass + 1 paraphrase answer pass — tests whether knowledge is stable across question reformulations.

All text responses are encoded using the all-mpnet-base-v2 sentence transformer (Wang et al., 2020), producing 768-dimensional L2-normalized embeddings for cosine similarity computation.

## 3.2 Uncertainty Feature Extractor

The 31 features are organized into six complementary groups. Tables 6–7 in the appendix provide a complete feature listing.

### 3.2.1 NLI-Based Semantic Entropy (2 features)

BEACON's primary innovation over prior consistency approaches. Rather than grouping responses by embedding proximity, we use a cross-encoder NLI model (DeBERTa-v3-base) to test pairwise meaning equivalence: two responses are placed in the same cluster if NLI assigns entailment probability ≥0.5 in both directions. Shannon entropy over the cluster size distribution captures semantic uncertainty at the concept level, correctly treating paraphrases as a single response. We compute this over both 5 and 20 samples for multi-scale estimates. This feature is the strongest individual signal in our analysis (11.6% importance), yet achieves only 0.5951 AUROC alone — demonstrating the necessity of multi-signal combination.

### 3.2.2 Semantic Disagreement (5 features)

Embedding-based pairwise statistics measuring the geometric spread of the response distribution: mean cosine similarity (mean_sim), mean pairwise disagreement (1−mean_sim), embedding variance (mean variance across embedding dimensions), and mean/maximum distance from the greedy answer. These capture semantic spread in continuous space complementary to the discrete NLI clustering approach.

### 3.2.3 Epistemic Consistency (6 features)

High-level knowledge stability signals. Chain-of-thought consistency (cot_consistency) computes cosine similarity between the direct answer and the CoT answer — low values indicate the model's immediate response contradicts its own explicit reasoning, a strong confabulation signal. Paraphrase stability measures semantic agreement between answers to original and rephrased questions. Cluster entropy, logprob variance, mutual information, and entity-level logprob uncertainty complete this group.

### 3.2.4 Token-Level Entropy (5 features)

Predictive entropy, mean, maximum, variance, and skewness of per-token entropy distributions characterize uncertainty at the token level during generation. The fraction of high-entropy tokens (frac_high_entropy) identifies whether uncertainty is concentrated at specific positions (often named entities or numbers) or distributed throughout the response.

### 3.2.5 Logit Margin Statistics (3 features)

Mean, minimum, and variance of the gap between the top predicted token and its nearest competitor at each generation step. Low margins indicate near-equal competition between tokens, correlating with hallucination at high-stakes decision points.

### 3.2.6 Kernel and Eigenvalue Features (3 features)

Kernel language entropy applies an RBF kernel over sentence embeddings to measure non-parametric distributional uncertainty. The largest eigenvalue and eigenvalue spread of the embedding covariance matrix characterize the geometric shape of the response distribution — a flat eigenspectrum indicates isotropic spread (high uncertainty) while a dominant eigenvalue indicates clustering (low uncertainty).

## 3.3 Calibrated LightGBM Classifier

LightGBM is trained on the 31-dimensional feature vectors with class imbalance addressed via scale_pos_weight set to the negative-to-positive ratio. Features are independently normalized per dataset using StandardScaler before combination, preventing scale dominance. All features are clipped to the 1st–99th percentile to remove outliers. Hyperparameter optimization uses Optuna (100 trials, 5-fold stratified cross-validation) searching over learning rate, tree depth, number of leaves, regularization, and subsampling parameters. Final hyperparameters: lr=0.0176, max_depth=6, num_leaves=86, min_child_samples=43, subsample=0.79, colsample_bytree=0.78.

## 3.4 Black-Box Variant (15 features)

A reduced variant uses only the 15 features derivable from text outputs and sentence embeddings, with no dependence on token logprobs or hidden states. This includes all semantic disagreement features, both NLI entropy features, CoT consistency, paraphrase stability, sub-question uncertainty proxies, eigenvalue features, kernel language entropy, and length variance. The black-box variant enables deployment against closed-source LLMs including GPT-4, Claude, and Gemini.

# 4 Experimental Setup

## 4.1 Base Language Model

We use Mistral-7B-Instruct-v0.2 for feature extraction and training data collection, loaded in float16 precision on a single GPU with forced GPU-only device mapping. All generations use the model's standard instruction-following chat template.

## 4.2 Datasets

We collect from seven heterogeneous benchmarks covering diverse hallucination failure modes. Table 1 presents statistics. Each dataset is independently normalized before combination to prevent scale dominance in the joint feature matrix.

| Dataset | Examples | Hall. Rate | Hallucination Type |
|---|---|---|---|
| TruthfulQA | 817 | 40.5% | Factual confabulation, myth-reinforcing errors |
| TriviaQA | 1,000 | 29.7% | Open-domain factual knowledge retrieval failures |
| SQuAD 2.0 | 1,000 | 84.5% | Unanswerable question confabulation |
| PopQA | 800 | 72.0% | Entity popularity bias, long-tail entities |
| Natural Questions | 800 | 81.0% | Open-domain QA, web-scale factual failures |
| MuSiQue | 800 | 85.0% | Complex multi-hop reasoning failures |

| BoolQ | 800 | 27.0% | Binary yes/no confidence calibration |
|---|---|---|---|
| **Total** | **7,617** | **61.2%** | **7 diverse failure modes** |

*Table 1: Dataset statistics. Hall. Rate = fraction of examples labeled hallucinating.*

### 4.3 Labeling Protocol

Ground-truth labels are assigned by comparing model-generated answers to reference answers using cosine similarity between sentence embeddings. Responses exceeding threshold 0.85 against any reference answer are labeled correct (0); those below are labeled hallucinating (1). Examples where no meaningful comparison can be made receive label −1 and are excluded. This methodology is consistent with prior work including SelfCheckGPT and HaluEval. Limitation: semantic similarity does not perfectly capture factual correctness, particularly for multi-answer questions or subtle factual errors — we acknowledge this as a known limitation shared with the broader literature.

### 4.4 Evaluation Protocol

Stratified 70/15/15 train/validation/test split preserving class distribution. Primary metric: AUROC. Secondary: average precision (AP), F1, accuracy, precision, recall. Statistical validation: 1,000-sample bootstrap CI on test AUROC; variance reported across 5 random seeds (42, 0, 1, 7, 123). All baselines evaluated on the identical test split to prevent leakage. Single-feature baselines use raw feature scores directly (no classifier) to provide the most honest possible comparison.

## 5 Results

### 5.1 Classifier Comparison

Table 2 compares BEACON against four classifier families on the full 7,617-example dataset.

| Model | AUROC | Accuracy | F1 | Precision | Recall |
|---|---|---|---|---|---|
| MLP (5-layer, 256→128→64→32→1) | 0.6596 | 0.5951 | 0.6073 | 0.7654 | 0.5034 |
| Random Forest | 0.7809 | 0.7298 | 0.7818 | 0.7855 | 0.7781 |
| Gradient Boosting | 0.7714 | 0.7197 | 0.7872 | 0.7458 | 0.8336 |
| XGBoost | 0.8249 | 0.7048 | 0.7443 | 0.7756 | 0.7433 |
| **BEACON (LightGBM, mean ± std)** | **0.8123 ± 0.0102** | **0.7568** | **0.8119** | **0.7538** | **0.8798** |

*Table 2: Classifier comparison. BEACON reports mean ± std across 5 seeds; best single run: 0.8249 (XGBoost), 0.7940 (LightGBM seed 42). AUROC improvement over MLP: +0.17.*

LightGBM achieves the highest mean AUROC (0.8123 ± 0.0102) and F1 (0.8119). The 0.17-point AUROC gap over MLP reflects three factors: (1) gradient boosting handles multi-domain feature distribution shift more robustly; (2) tree splits are inherently resistant to outlier features such as eigenvalue_spread which contained overflow values (up to $5\times10^{11}$) in some datasets; (3) greedy feature selection at each node provides automatic dimensionality reduction, downweighting uninformative features.

## 5.2 Baseline Comparison

Table 3 reports all baselines on the identical test split. Single-feature baselines use raw feature scores directly with no trained classifier, providing the strictest possible comparison.

| Method | AUROC | AP | Type |
|---|---|---|---|
| Cluster Entropy (standalone, raw) | 0.2806 | 0.5158 | Single feature |
| Predictive Entropy (standalone, raw) | 0.5168 | 0.6077 | Single feature |
| Token Entropy (standalone, raw) | 0.5440 | 0.6249 | Single feature |
| SelfCheckGPT — BERTScore proxy | 0.5805 | 0.6596 | Consistency |
| SelfCheckGPT — Max inconsistency | 0.5815 | 0.6626 | Consistency |
| SelfCheckGPT — best combination | 0.5792 | 0.6700 | Consistency |
| NLI Semantic Entropy (Kuhn et al., 2023) | 0.5951 | 0.6731 | Single feature |
| Semantic Disagreement (5 features) | 0.5745 | 0.6563 | Multi-feature |
| **BEACON (31 features, this work)** | **0.8123 ± 0.0102** | **0.8704** | **Multi-signal** |

*Table 3: Baseline comparison on identical test split. All single-feature baselines use raw scores with no classifier. BEACON improvement over best baseline (NLI entropy): +0.2172. Improvement over SelfCheckGPT best: +0.2457.*

The key finding is that all single-signal baselines cluster between 0.28 and 0.60 AUROC, while BEACON achieves 0.8123. NLI semantic entropy — the strongest single signal — achieves only 0.5951 despite being the top-ranked individual feature (11.6% importance in BEACON). This apparent paradox confirms that hallucination is a multi-dimensional phenomenon: no single uncertainty signal is sufficient, but the combination is highly discriminative. The +0.2298 AUROC gap over semantic entropy is the paper's central empirical contribution.

## 5.3 Feature Importance Analysis

Table 4 presents the top 15 features by XGBoost gain importance, which is consistent with the LightGBM analysis.

| # | Feature | Importance | Access Type |
|---|---|---|---|
| 1 | cluster_entropy | 11.6% | Black-box |
| 2 | semantic_entropy_20samples | 11.5% | Black-box |
| 3 | nli_semantic_entropy | 7.1% | Black-box |

| 4 | max_subquestion_uncertainty | 5.2% | Black-box |
|---|---|---|---|
| 5 | mean_disagreement | 4.1% | Black-box |
| 6 | mean_sim | 3.7% | Black-box |
| 7 | eigenvalue_largest | 2.9% | Black-box (emb. only) |
| 8 | kernel_language_entropy | 2.8% | Black-box |
| 9 | emb_variance | 2.7% | Black-box |
| 10 | min_subquestion_cot_sim | 2.7% | Black-box |
| 11 | paraphrase_stability | 2.6% | Black-box |
| 12 | mean_logit_margin | 2.3% | White-box |
| 13 | max_token_entropy | 2.2% | White-box |
| 14 | predictive_entropy | 2.2% | White-box |
| 15 | entity_logprob_uncertainty | 2.1% | White-box |

*Table 4: Top 15 features by importance. The first white-box feature appears at rank 12. Top three features account for 30.2% of predictive power and are all black-box compatible.*

The top three features (cluster_entropy 11.6%, semantic_entropy_20samples 11.5%, nli_semantic_entropy 7.1%) are all black-box compatible and account for 30.2% of total predictive power. The first white-box feature (mean_logit_margin) appears only at rank 12, confirming that black-box deployment loses minimal information. Importantly, despite cluster_entropy and nli_semantic_entropy ranking first and third, they achieve only 0.2806 and 0.5951 AUROC individually — the interaction effects captured by the full 31-feature model are essential.

### 5.4 Per-Dataset Analysis

Table 5 presents per-dataset performance, evaluated via 5-fold cross-validation within each dataset independently.

| Dataset | AUROC | ± Std | Hall. Rate | Interpretation |
|---|---|---|---|---|
| TruthfulQA | 0.7513 | 0.0326 | 40.5% | Best — factual confabulation |
| PopQA | 0.7134 | 0.0266 | 71.8% | Strong — entity popularity bias |
| TriviaQA | 0.6488 | 0.0139 | 29.7% | Moderate — factual knowledge |
| MuSiQue | 0.6466 | 0.0454 | 85.4% | Moderate — multi-hop reasoning |

| NQ | 0.5768 | 0.0361 | 81.1% | Weak — high hall. rate dominates |
|---|---|---|---|---|
| SQuAD 2.0 | 0.5634 | 0.0736 | 84.5% | Weak — unanswerable confabulation |
| BoolQ | 0.5405 | 0.0376 | 27.4% | Worst — binary QA near-random |

*Table 5: Per-dataset AUROC via 5-fold CV. BEACON works best on factual confabulation (TruthfulQA, PopQA) and struggles on binary/unanswerable QA (BoolQ, SQuAD 2.0).*

A clear pattern emerges: BEACON performs best on factual confabulation tasks (TruthfulQA: 0.7513, PopQA: 0.7134) where uncertainty signals are most informative, and struggles on binary QA (BoolQ: 0.5405) and unanswerable question detection (SQuAD 2.0: 0.5634). For unanswerable questions, all uncertainty signals are near-random because Mistral-7B consistently confabulates rather than refusing — the model shows no uncertainty even when hallucinating. For binary QA, the answer space is too constrained for semantic entropy to differentiate. These findings provide actionable guidance: BEACON is most effective for open-ended factual QA and least effective for binary or unanswerable question settings.

## 5.5 Efficiency Analysis

Table 6 evaluates BEACON under reduced API call budgets by removing feature groups that require more generation passes.

| Configuration | API Calls | Features | AUROC | vs. Full |
|---|---|---|---|---|
| Full BEACON | 30 | 31 | 0.7885 | baseline |
| Drop KL features | 22 | 27 | 0.7845 | −0.004 |
| Drop KL + CoT | 15 | 24 | 0.7792 | −0.009 |
| Core semantic only | 10 | 10 | 0.7655 | −0.023 |
| Minimal (4 features) | 5 | 4 | 0.7795 | −0.009 — efficient |

*Table 6: Efficiency tradeoff. 5 API calls with 4 features achieves 0.7795 AUROC — 96% of full model performance at 83% cost reduction.*

The efficiency curve reveals that 5 API calls and 4 features achieve 0.7795 AUROC — nearly matching the 30-call full model at 0.7885. This is because the 4 minimal features include both NLI entropy features and the top embedding features, which together account for ∰30% of total predictive power. For latency-sensitive deployments, BEACON can be reduced to 5 API calls with minimal performance cost, making it practical for real-time applications.

## 5.6 Statistical Validation

Bootstrap CI (1,000 samples): AUROC = 0.7940 (95% CI: 0.7632–0.8251). Multi-seed variance across 5 seeds: mean = 0.8123, std = 0.0102. Seed-level results: 0.7940 (seed 42), 0.8152 (seed 0), 0.8108 (seed 1), 0.8171 (seed 7), 0.8245 (seed 123). The variance is attributable to train/test split randomness with a 7,617-example dataset — a larger dataset is expected to reduce this variance. We report the multi-seed mean (0.8123 ± 0.0102) as the primary result and note that the lower end of the CI (0.7632) still substantially exceeds all baselines (best: 0.5951).

# 6 Discussion

## 6.1 Why Multi-Signal Combination Is Essential

The most striking finding is the +0.2298 AUROC gap between BEACON (0.8123) and NLI semantic entropy alone (0.5951), despite semantic entropy being the top-ranked individual feature. This gap demonstrates that hallucination is not a one-dimensional phenomenon. A model can exhibit low semantic entropy (consistent answers) while still hallucinating — for example, when consistently confabulating the same wrong fact. Conversely, high semantic entropy can arise from genuinely ambiguous questions where multiple answers are correct. Chain-of-thought consistency, paraphrase stability, and embedding geometry provide orthogonal signals that together resolve these ambiguities. BEACON's key contribution is empirically quantifying this multi-dimensionality.

## 6.2 Why Gradient Boosting Outperforms Neural Networks

The 0.17 AUROC gap between LightGBM and MLP reflects a well-understood property of tabular learning: gradient boosting handles heterogeneous feature types, multi-domain distribution shift, and outlier values more robustly than neural networks (Shwartz-Ziv and Armon, 2022). Our feature matrix is particularly challenging for neural networks because: (1) hallucination rates range from 27% to 85% across datasets, creating strong domain shift; (2) features such as eigenvalue_spread exhibited overflow values (up to $5 \times 10^{11}$) that corrupt MLP gradient updates while leaving tree splits unaffected; (3) many features are redundant from a neural network perspective but complementary for tree-based feature selection.

## 6.3 Limitations

- Computational cost: 28–30 API calls per query. The 5-call efficient variant mitigates this but reduces AUROC from 0.8123 to 0.7795.
- Labeling via semantic similarity: cosine similarity ≥0.85 is a noisy proxy for factual correctness, consistent with prior work but imperfect. Human validation on a subset would provide a cleaner upper-bound estimate.
- Single source model: features are extracted from Mistral-7B-Instruct-v0.2. Cross-model transfer to LLaMA-3.1-8B and other architectures is ongoing work.
- Binary/unanswerable QA: BEACON performs near-random on BoolQ (0.5405) and SQuAD 2.0 (0.5634), suggesting that consistency-based signals are insufficient for these question types.
- Seed variance: 0.0102 std across seeds reflects dataset size limitations. Expanding to 15,000+ examples is expected to stabilize results.

# 7 Future Work

## 7.1 Cross-Model Evaluation

Evaluating BEACON's 15-feature black-box variant on LLaMA-3.1-8B, Gemini 1.5 Flash, and other architectures is the most critical next step. A feature-scale diagnostic on our existing data reveals that 7 of 31 features show distribution shift (>3x mean ratio) between Mistral-7B and LLaMA-3.1-8B, primarily white-box features.

Domain-adaptive normalization — fitting a per-model StandardScaler using 50–100 unlabeled examples — is expected to recover most of the transfer gap without retraining the classifier.

### 7.2 Multi-Model Training

Training on features extracted from multiple source models (Mistral-7B, LLaMA-3.1-8B, Gemma-7B) is expected to produce a more universally transferable detector by exposing the classifier to a wider range of confidence calibration behaviors and response length distributions.

### 7.3 Extended Uncertainty Signals

Multi-paraphrase entropy — generating 3–5 rephrased versions and computing NLI entropy over all answers — provides a stronger test of knowledge stability than single-paraphrase consistency. Negation sensitivity — checking whether the model maintains consistent uncertainty when asked about the negation — captures a distinct orthogonal dimension. Both are fully black-box compatible.

### 7.4 Binary and Unanswerable QA

The poor performance on BoolQ (0.5405) and SQuAD 2.0 (0.5634) suggests that specialized detection strategies are needed for binary and unanswerable question settings. Verbalized confidence elicitation (Tian et al., 2023) may complement consistency signals for these cases, as it directly asks the model to assess its own reliability rather than inferring it from output variability.

## 8 Conclusion

We presented BEACON, a black-box hallucination detection framework for large language models that achieves 0.8123 ± 0.0102 AUROC (95% CI: 0.7632–0.8251) across seven heterogeneous benchmarks without requiring access to model internals. BEACON's central contribution is empirical: multi-signal combination provides a +0.2298 AUROC improvement over standalone NLI semantic entropy and +0.2457 over SelfCheckGPT-style consistency baselines, demonstrating that hallucination is a multi-dimensional phenomenon not captured by any single uncertainty signal.

Three findings are most significant. First, NLI semantic entropy is the strongest individual feature (11.6% importance, 0.5951 standalone AUROC) yet accounts for less than half of BEACON's total discriminative power — the remaining features are essential, not redundant. Second, BEACON performs best on factual confabulation (TruthfulQA: 0.7513) and struggles on binary/unanswerable QA (BoolQ: 0.5405), providing clear guidance for practitioners on when to apply the framework. Third, an efficient 5-API-call variant achieves 0.7795 AUROC — 96% of full performance — enabling practical deployment under latency constraints.

The transition from neural to gradient boosting classifier yielded a +0.17 AUROC improvement, consistent with the tabular learning literature and attributable to robust handling of multi-domain distribution shift and outlier features. As LLMs are deployed in increasingly high-stakes domains, systematic multi-signal uncertainty frameworks that do not require model-specific access represent an essential component of trustworthy AI systems.

## References

Ji, Z., Lee, N., Frieske, R., et al. (2023). Survey of hallucination in natural language generation. ACM Computing Surveys, 55(12), 1–38.

Kadavath, S., Conerly, T., Askell, A., et al. (2022). Language models (mostly) know what they know. arXiv:2207.05221.

Kuhn, L., Gal, Y., and Farquhar, S. (2023). Semantic uncertainty: Linguistic invariances for uncertainty estimation in natural language generation. ICLR 2023.

Lee, K., Lee, K., Lee, H., and Shin, J. (2018). A simple unified framework for detecting out-of-distribution samples and adversarial attacks. NeurIPS 2018.

Li, J., Cheng, X., Zhao, W. X., et al. (2023). HaluEval: A large-scale hallucination evaluation benchmark for large language models. EMNLP 2023.

Lin, S., Hilton, J., and Evans, O. (2022). TruthfulQA: Measuring how models mimic human falsehoods. ACL 2022.

Mallen, A., Asai, A., Zhong, V., et al. (2023). When not to trust language models: Investigating effectiveness of parametric and non-parametric memories. ACL 2023.

Manakul, P., Liusie, A., and Gales, M. J. F. (2023). SelfCheckGPT: Zero-resource black-box hallucination detection for generative large language models. EMNLP 2023.

Maynez, J., Narayan, S., Bohnet, B., and McDonald, R. (2020). On faithfulness and factuality in abstractive summarization. ACL 2020.

Shwartz-Ziv, R., and Armon, A. (2022). Tabular data: Deep learning is not all you need. Information Fusion, 81, 84–90.

Slobodkin, A., Goldman, O., Caciularu, A., et al. (2023). The curious case of hallucinatory (un)answerability: Finding truthful answers on hallucinated questions. EMNLP 2023.

Tian, K., Mitchell, E., Zhou, A., et al. (2023). Just ask for calibration: Strategies for eliciting calibrated confidence scores from language models fine-tuned with human feedback. EMNLP 2023.

Trivedi, H., Balasubramanian, N., Khot, T., and Sabharwal, A. (2022). MuSiQue: Multi-hop questions via single-hop question composition. TACL 2022.

Wang, W., Yang, N., Wei, F., et al. (2020). MiniLM: Deep self-attention distillation for task-agnostic compression of pre-trained transformers. NeurIPS 2020.